\documentclass[letterpaper, 10pt, conference]{ieeeconf}

\IEEEoverridecommandlockouts                              
\usepackage{graphicx}
\usepackage{booktabs}
\usepackage{amsmath}
\usepackage{amssymb}
\DeclareMathSizes{10}{10}{8}{7}
\usepackage{diagbox}
\usepackage{multirow}
\usepackage[table,dvipsnames]{xcolor}
\usepackage{subcaption}
\usepackage{hyperref}
\hypersetup{
    colorlinks=true,
    linkcolor=blue,
    urlcolor=magenta
    }

\makeatletter
\def\blfootnote{\gdef\@thefnmark{}\@footnotetext}
\makeatother

\usepackage{textcomp}
\usepackage{color,colortbl}
\definecolor{beaublue}{rgb}{0.74, 0.83, 0.9}

\usepackage{xfp}
\usepackage{siunitx}
\newcommand{\avgseedstats}[8]{%
  \edef\avgmeanraw{\fpeval{(#1+#2+#3+#4)/4}}%
  \edef\avgseedraw{\fpeval{sqrt((#5^2+#6^2+#7^2+#8^2)/4)}}%
}
\newcommand{\avgseed}[8]{{\avgseedstats{#1}{#2}{#3}{#4}{#5}{#6}{#7}{#8}%
  \num[round-precision=2,round-mode=places]{\avgmeanraw} $\pm$ \num[round-precision=2,round-mode=places]{\avgseedraw}}}
\newcommand{\avgseedm}[8]{{\avgseedstats{#1}{#2}{#3}{#4}{#5}{#6}{#7}{#8}%
  \num[round-precision=2,round-mode=places]{\avgmeanraw} \pm \num[round-precision=2,round-mode=places]{\avgseedraw}}}

\title{\LARGE \bf
Enhancing Human-Likeness in Reinforcement Learning Agents via Hierarchical Macro Action Quantization
}

\author{M. Shaheer Luqman$^\dagger$~~~~~Usman Nizamani$^\dagger$~~~~~Fawad Javed Fateh~~~~~Ali Shah Ali~~~~~Murad Popattia\\Quoc-Huy Tran~~~~~M. Zeeshan Zia
   \thanks{All authors are with Retrocausal, Inc., Redmond, WA 98052, USA.\newline
	Website: \url{www.retrocausal.ai}\newline
	Email: {\tt\small \{shaheer,usman,fawad,alishah,murad,\newline
	huy,zeeshan\}@retrocausal.ai}}}

\begin{document}

\maketitle
\thispagestyle{empty}
\pagestyle{empty}


\begin{abstract}
Human-like agents are a long-standing goal of artificial intelligence. Despite strong performance, most reinforcement learning (RL) agents remain reward-driven and often exhibit behaviors that differ from humans, limiting interpretability and reliability. In this work, we introduce a novel human-like RL framework that predicts action sequences closely aligned with human behaviors while maximizing rewards. Specifically, we encode human demonstrations into macro actions using a hierarchical macro action quantization approach (HiMAQ) consisting of two successive levels of vector quantization. The lower quantization level maps input actions to fine-grained subaction clusters, while the higher quantization level aggregates these subaction clusters into action clusters. Extensive evaluations on the D4RL benchmarks show that our hierarchical approach outperforms the non-hierarchical baseline (MAQ), achieving higher human-likeness scores and better success rates than previous RL agents. The improvements generalize across integrations with various RL algorithms, namely IQL, SAC, and RLPD.
\end{abstract}


\section{Introduction}
\label{sec:introduction}

As robots take on increasingly complex tasks alongside humans~\cite{schrittwieser_mastering_2020, vinyals_grandmaster_2019, openai_dota_2019, hafner_mastering_2025, fu_d4rl_2021, ball_efficient_2023}, a key expectation emerges: motion that is not merely successful, but recognizably \textit{human-like}.
Conventional reward-driven RL policies routinely produce unnatural, jerky motions~\cite{mysore_regularizing_2021, lee_gradientbased_2024, shen_deep_2020, koch_flight_2019, milani_navigates_2023, devlin_navigation_2021, zuniga_how_2022, ho_humanlike_2024} that erode trust and fail to transfer the rich knowledge encoded in human demonstrations. 

Recent work has begun to close this gap by grounding RL within the space of human demonstrations.
Human-like receding horizon control (HRC)~\cite{hansen2022temporal} constrains trajectory search to human-generated segments, producing more structured motion.
MAQ~\cite{guo2026learning} extends this by encoding human trajectory segments into a discrete codebook via a conditional vector quantized variational autoencoder (VQ-VAE)~\cite{van2017neural}, allowing a policy to select from a compact set of learned macro actions. However, MAQ's single-level codebook treats each macro action as a single atomic prototype, limiting its ability to capture the full variation within each motion segment. A reaching motion, for instance, spans a broad approach, fine grasp alignment, and controlled contact --- collapsing all of this into one prototype loses the subtle dynamics that make motion feel human.

We propose \textbf{Hi}erarchical \textbf{M}acro \textbf{A}ction \textbf{Q}uantization (HiMAQ), inspired by MAQ~\cite{guo2026learning} and hierarchical VQ-VAE~\cite{spurio2025hierarchical}, a two-level vector quantization framework that captures human demonstrations at both macro action and fine-grained subaction levels.
The key insight is that instead of quantizing an encoded trajectory directly to an action prototype, HiMAQ first maps it to a fine-grained \textit{subaction} prototype, which is then mapped to a coarser \textit{action} prototype. Fitting the action prototypes to the subaction structure, rather than to the encoded trajectories directly, places them so that the same number of prototypes covers the demonstrated behavior more closely. Evaluated on four dexterous manipulation tasks from D4RL's Adroit benchmark~\cite{fu_d4rl_2021} across three RL algorithms (IQL~\cite{kostrikov_offline_2021}, SAC~\cite{haarnoja_soft_2018}, and RLPD~\cite{ball_efficient_2023}), HiMAQ outperforms MAQ~\cite{guo2026learning} on trajectory similarity metrics, while yielding better success rates. In a human study with 22 evaluators, HiMAQ+RLPD achieves a Turing test win rate of 43\%, and HiMAQ agents outperform all competing methods in the human-likeness ranking test.

In summary, our contributions are:
\begin{itemize}
    \item \textit{HiMAQ}, a hierarchical macro action quantization framework that encodes human demonstrations through two levels of vector quantization --- subaction prototypes capturing fine-grained structure, and action prototypes capturing coarse motion patterns --- enabling richer human motion representation than single-level approaches.

    \item \textit{Consistent human-likeness gains across RL algorithms.} Averaged over four Adroit tasks, HiMAQ improves trajectory-similarity metrics (DTW and Wasserstein distance, on both states and actions) over MAQ~\cite{guo2026learning} when combined with IQL~\cite{kostrikov_offline_2021}, SAC~\cite{haarnoja_soft_2018}, and RLPD~\cite{ball_efficient_2023}, while improving task success rates.

    \item \textit{Perceptual validation via human study.} With 22 human evaluators across Turing and ranking tests, HiMAQ variants outperform MAQ~\cite{guo2026learning} counterparts, confirming that quantitative trajectory improvements translate to perceptually recognizable human-likeness.
\end{itemize}

\section{Related Work}
\label{sec:relatedwork}

\noindent \textbf{Human-Likeness in Reinforcement Learning.}
RL often prioritizes reward optimization over realistic behavior generation. Human-like RL approaches can be classified into two groups. One line of work introduces constraints or penalties to suppress unnatural behaviors, e.g., Fujii et al.~\cite{fujii2013evaluating} penalize non-human actions, while Ho et al.~\cite{ho2024towards} design penalties for unrealistic movements. Despite great performance, the above methods rely heavily on manually designed rules. Another direction learns from human demonstrations collected in domains such as gaming~\cite{hester2018deep}, autonomous driving~\cite{bojarski2016end}, and robotics~\cite{fu_d4rl_2021}. Techniques such as behavior cloning and inverse RL~\cite{zare2024survey,arora2021survey} can transfer human behavioral traits to agents, but are often limited by dataset quality and diversity. Recent work models behavior through learned macro actions, e.g., MAQ~\cite{guo2026learning} learns discrete actions from human trajectories using VQ-VAE~\cite{van2017neural}. In this work, we propose a hierarchical macro action quantization framework, yielding higher human-likeness and better task performance. Diffusion-based imitation~\cite{pearce2023imitating} and Flow Q-Learning~\cite{park2025flow} offer alternatives for human behavior modeling and offline-to-online RL. We compare HiMAQ with MAQ under identical RL algorithms to isolate hierarchical quantization’s contribution, leaving generative-policy comparisons to future work.

\noindent \textbf{Macro Action Representation Learning.} To enable temporally extended decision-making, MAQ~\cite{guo2026learning} adopts Semi-Markov Decision Processes (SMDPs). Formally, at state $\textbf{s}_t$, the agent executes a macro action $\mathbf{m}_t=(\mathbf{a}_t,\mathbf{a}_{t+1},\dots,\mathbf{a}_{t+H-1})$ over $H$ time steps. SMDP is defined by $(\mathcal{S},\mathcal{M},R,P,\gamma)$, where $\mathcal{S}$ is the state space and $\mathcal{M}$ is the set of macro actions. Executing $\textbf{m}_t$ from state $\textbf{s}_t$ yields an aggregate reward across the multi-step execution interval, specified as $R(\mathbf{s}_t,\mathbf{m}_t)=\sum_{k=t}^{t+H-1} r(\mathbf{s}_k,\mathbf{a}_k)$~\cite{durugkar2016deep}. The policy objective is to maximize the expected discounted return over macro actions: $J(\pi_\theta)=\mathbb{E}_{(\mathbf{s}_t,\mathbf{m}_t)\sim\pi_\theta}\left[\sum_{t=0}^{\infty}\gamma^t R(\mathbf{s}_t,\mathbf{m}_t)\right]$. By replacing fine-grained actions with macro actions, SMDPs enable long-horizon planning. This reduces decision frequency while preserving long-term behavioral structure. Our work follows this line of research.

\noindent \textbf{Vector Quantized Variational Autoencoders.} Building on the classical variational autoencoder (VAE)~\cite{kingma2013auto}, VQ-VAE~\cite{van2017neural} introduces a discrete latent codebook. An encoder maps inputs into continuous embeddings, which are quantized to the nearest codebook vectors, while a decoder reconstructs the input from these discrete tokens. Such representations yield compact, structured features suitable for sequential modeling and reinforcement learning. For example, Ozair et al.~\cite{ozair2021vector} and Antonoglou et al.~\cite{antonoglou2021planning} deploy conditional VQ-VAEs to model stochastic environments, while Luo et al.~\cite{luo2023action} project continuous action spaces into discrete, reusable action primitives to facilitate policy optimization. Hierarchical variants further capture behaviors at multiple abstraction levels; for example, Spurio et al.~\cite{spurio2025hierarchical} separate low-level subactions from high-level strategies. Inspired by the success of~\cite{spurio2025hierarchical} in temporal action segmentation, we develop a hierarchical macro action quantization framework to improve human-likeness in RL agents.
\section{Our Approach}
\label{sec:method}

\subsection{Human-Like Trajectory Optimization}

Robot manipulation tasks require agents to generate smooth, human-like trajectories by learning from human demonstrations, where trajectory optimization provides a principled framework for this purpose.\ Trajectory optimization is defined under a Markov Decision Process (MDP) $(\mathcal{S}, \mathcal{A}, \mathcal{P}, R, \gamma, T, p_0)$, where $\mathcal{S}$ and $\mathcal{A}$ denote the continuous state and action spaces, respectively. The transition dynamics are given by the policy $\mathcal{P} : \mathcal{S} \times \mathcal{A} \rightarrow \Delta(\mathcal{S})$, with reward function $R : \mathcal{S} \times \mathcal{A} \rightarrow \mathbb{R}$, discount factor $\gamma \in [0,1)$, episode length $T$, and initial state distribution $p_0$. The goal is to find an action sequence that maximizes the expected cumulative discounted return over the trajectory. The human-like receding horizon control (HRC) framework~\cite{hansen2022temporal} is used for trajectory optimization over sequences of actions, where at each timestep $t$, a sequence of $H$ actions ($H$ is much smaller than $T$) is optimized jointly. Replanning is performed every $j$ steps, with $1 \leq j \leq H$, after executing the first $j$ actions of $\textbf{a}^*_{t:t+H-1}$. Human-likeness is enforced by restricting the search space to human-generated trajectory segments $\mathcal{H}$ sampled from a dataset $\mathcal{D} = \{\tau^{(i)}\}_{i=1}^{N}$:
\begin{multline}
\textbf{a}_{t:t+H-1}^{*}
=
{\arg\max}_{\textbf{a}_{t:t+H-1}\in\mathcal{H}}
\mathbb{E}
\bigg[
\sum_{i=t}^{t+H-1}\gamma^{i-t}R(\textbf{s}_i,\textbf{a}_i)
\\
\Big\rvert
\textbf{s}_t;
\textbf{a}_{t:t+H-1}
\bigg].
\label{eq:HRC}
\end{multline}
Replanning after every $j$ steps is computationally expensive due to repeated searching over all human trajectory segments. Furthermore, the current state $\textbf{s}_t$ rarely exactly matches the starting state of stored demonstration segments, making direct reuse of continuous action trajectories unreliable. MAQ~\cite{guo2026learning} addresses these limitations by representing behavior as fixed-length $H$-step action sequences $(\textbf{a}_t, \textbf{a}_{t+1}, \ldots, \textbf{a}_{t+H-1})$, which are encoded into discrete latent codes using a conditional VQ-VAE~\cite{van2017neural}. Given a macro action segment, the encoder maps the trajectory into a latent embedding, which is then quantized to the nearest codebook vector $\textbf{e}_t$. The corresponding discrete code, together with the current state $\textbf{s}_t$, is passed to the decoder to reconstruct the macro action sequence $\widetilde{\textbf{m}}_t$. Despite its effectiveness, MAQ~\cite{guo2026learning} quantizes entire $H$-step trajectories using a single-level codebook, limiting its ability to capture the fine-grained structure of complex macro actions. HiMAQ addresses this limitation through subaction quantization and segmentation, which captures subtle motion variations within each macro action to improve action quantization and segmentation.

\begin{figure*}[t]
	\centering
		\includegraphics[width=0.85\linewidth, trim = 0mm 10mm 0mm 0mm, clip]{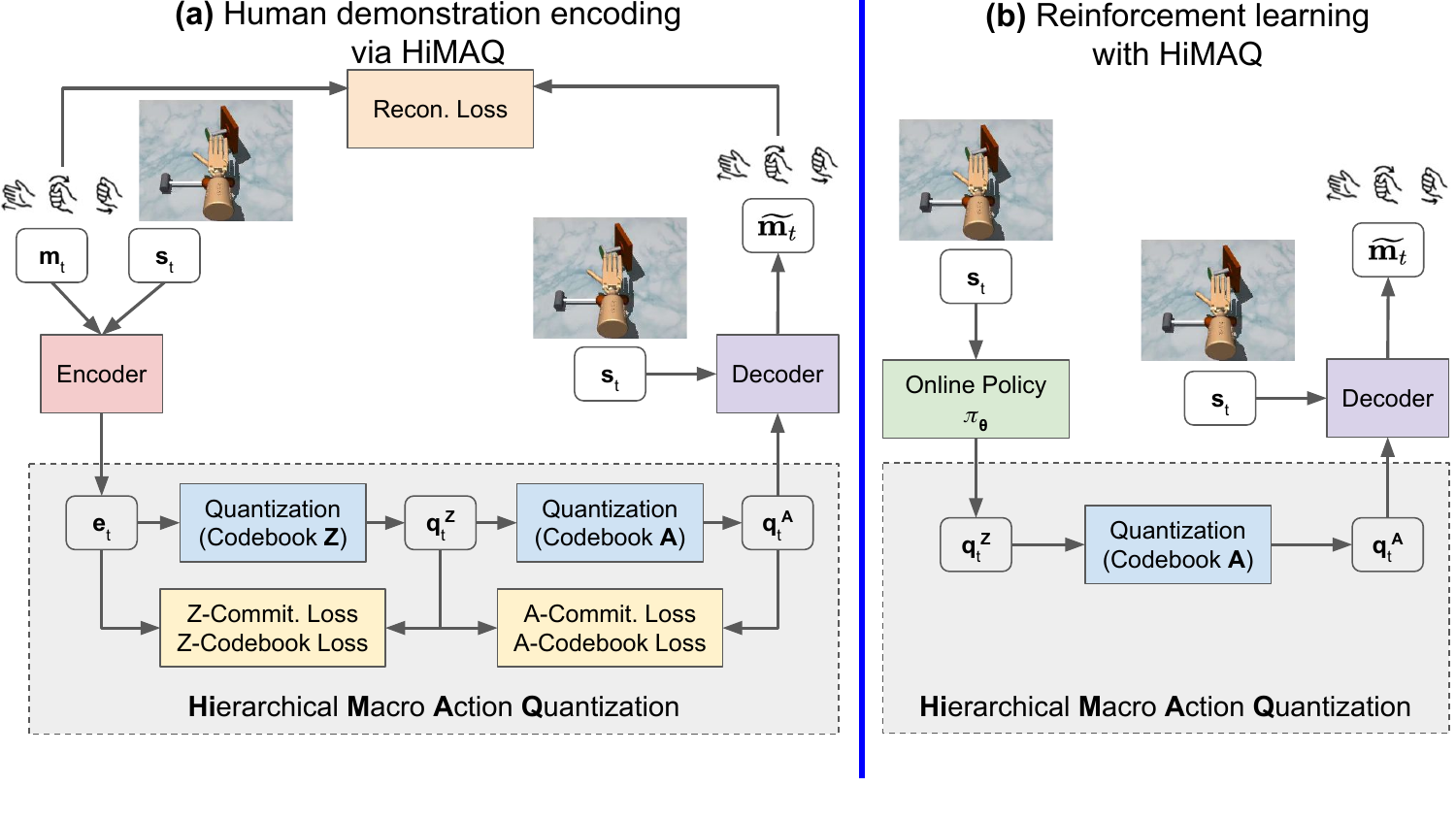}
	\caption{Our HiMAQ framework includes a two-stage training process: \textit{(a) Human demonstration encoding via HiMAQ.} The state ($\textbf{s}_t$) and macro action ($\textbf{m}_t =\{\textbf{a}_t, \textbf{a}_{t+1}, \dots, \textbf{a}_{t+H-1} \}$) are embedded via an encoder into the latent vector $\textbf{e}_t$. Next, $\textbf{e}_t$ is mapped to the nearest subaction prototype $\textbf{q}^\textbf{Z}_t$, which is then mapped to the closest action prototype $\textbf{q}^\textbf{A}_t$. Lastly, $\textbf{q}^\textbf{A}_t$ and $\textbf{s}_t$ are then passed to a decoder, which yields the predicted macro action $\widetilde{\textbf{m}}_t$. We train the encoder, decoder and both codebooks by using a combination of commitment, codebook, and reconstruction losses. \textit{(b) Reinforcement learning with HiMAQ.} We train an online policy $\pi_{\theta}$, which takes $\textbf{s}_t$ as input and predicts the subaction prototype $\textbf{q}^\textbf{Z}_t$. Next, $\textbf{q}^\textbf{A}_t$ is derived as the closest action prototype to $\textbf{q}^\textbf{Z}_t$. Finally, $\textbf{q}^\textbf{A}_t$ and $\textbf{s}_t$ are fed into the decoder to produce the predicted macro action $\widetilde{\textbf{m}}_t$.}
	\label{fig:method}
\end{figure*}

\subsection{Human Demonstration Encoding with HiMAQ}
Inspired by the success of HVQ~\cite{spurio2025hierarchical} in temporal action segmentation, HiMAQ represents macro actions from offline 
human demonstrations using a two-level hierarchical codebook structure, 
$\textbf{Z} = \{\textbf{z}_j\}_{j=1}^{\alpha K}$ for subactions and $\textbf{A} = \{\textbf{a}_i\}_{i=1}^{K}$ 
for actions, as illustrated in Fig.~\ref{fig:method}(a). Here, $K$ denotes the number of action codebook entries and $\alpha$ is a ratio 
parameter controlling the size of the subaction codebook, where $\textbf{A}$ 
contains $K$ action prototypes and $\textbf{Z}$ contains $\alpha K$ subaction 
prototypes. The encoder takes the state $\textbf{s}_t$ and macro action
$\textbf{a}_{t:t+H-1}$ as input and generates the latent embedding $\textbf{e}_t$.
Both the encoder and decoder are implemented as two-layer MLPs, and both codebooks $\textbf{Z}$ and $\textbf{A}$ are initialized with random samples from a standard normal distribution; all other architectural choices (hidden sizes, latent dimension) are kept identical to MAQ~\cite{guo2026learning} to isolate the effect of hierarchical quantization.

The first quantization level maps the encoder embedding $\textbf{e}_t$ to its 
nearest prototype $\textbf{z}_{j^*} \in \textbf{Z}$ using Euclidean distance, producing 
the quantized subaction representation $\textbf{q}^\textbf{Z}_t$:
\begin{equation}
\textbf{q}^\textbf{Z}_t = \textbf{z}_{j^*}, \quad j^* = \arg\min_{j} \|\textbf{e}_t - \textbf{z}_j\|^2.
\end{equation}
The subaction representation $\textbf{q}^\textbf{Z}_t$ is then passed to the 
second level, where the nearest prototype $\textbf{a}_{i^*} \in \textbf{A}$ 
is computed using Euclidean distance, producing the action representation 
$\textbf{q}^\textbf{A}_t$:
\begin{equation}
\textbf{q}^\textbf{A}_t = \textbf{a}_{i^*}, \quad i^* = \arg\min_{i} \|\textbf{q}^\textbf{Z}_t - \textbf{a}_i\|^2.
\end{equation}
Finally, $\textbf{q}^\textbf{A}_t$ along with the state $\textbf{s}_t$ is passed to the decoder 
to reconstruct the macro action $\widetilde{\textbf{m}}_t$. Subaction quantization/segmentation models subtle motion variations within each macro action, improving action quantization/segmentation accuracy.

To train HiMAQ, we use a combination of reconstruction loss, commitment losses, and codebook losses, defined as:
\begin{equation}
\mathcal{L}_{recon} = \|\textbf{m}_t - \widetilde{\textbf{m}}_t\|^2,
\end{equation}
\begin{align}
\mathcal{L}_{commit_Z} &= \|\textbf{e}_t - sg[\textbf{q}^\textbf{Z}_t]\|^2, \notag \\
\mathcal{L}_{commit_A} &= \|\textbf{q}^\textbf{Z}_t - sg[\textbf{q}^\textbf{A}_t]\|^2,
\end{align}
\begin{align}
\mathcal{L}_{codebook_Z} &= \|sg[\textbf{e}_t] - \textbf{q}^\textbf{Z}_t\|^2, \notag \\
\mathcal{L}_{codebook_A} &= \|sg[\textbf{q}^\textbf{Z}_t] - \textbf{q}^\textbf{A}_t\|^2.
\end{align}
Here, the reconstruction loss $\mathcal{L}_{recon}$ measures the discrepancy between the original macro action $\textbf{m}_t$ and the reconstructed macro action $\widetilde{\textbf{m}}_t$.
The codebook losses fit each level to the population it quantizes: $\mathcal{L}_{codebook_Z}$ moves each subaction prototype towards the encoder embeddings assigned to it, while $\mathcal{L}_{codebook_A}$ moves each action prototype towards the subaction prototypes assigned to it, so that $\textbf{A}$ summarizes $\textbf{Z}$ rather than $\{\textbf{e}_t\}$ directly.
The commitment losses regularize the input of each quantization step: $\mathcal{L}_{commit_Z}$ keeps the encoder output $\textbf{e}_t$ close to $\textbf{Z}$, as in the standard VQ-VAE~\cite{van2017neural}, and $\mathcal{L}_{commit_A}$ keeps $\textbf{q}^\textbf{Z}_t$ close to $\textbf{A}$, so that both levels share a common embedding space. Also, 
$sg[\cdot]$ denotes the stop-gradient operator. A codebook loss updates only codebook parameters, while a commitment loss updates only the encoder or the preceding quantization level.

The total training loss combines the reconstruction loss $\mathcal{L}_{recon}$, 
the commitment losses, and the codebook losses, 
across both quantization levels as:
\begin{multline}
\mathcal{L} = \mathcal{L}_{recon} + \lambda_{commit}(\mathcal{L}_{commit_Z} +
\mathcal{L}_{commit_A}) \\+ \lambda_{codebook}(\mathcal{L}_{codebook_Z} + \mathcal{L}_{codebook_A}).
\end{multline}
For our experiments, we set $\lambda_{commit} = 0.25$ and $\lambda_{codebook} = 1.0$, following MAQ~\cite{guo2026learning}. As we will show in Sec.~\ref{sec:experiments}, HiMAQ achieves higher human-likeness scores and better success rates than prior works. Imposing trajectory smoothness (e.g., Lipschitz regularization~\cite{vuong2025action}) may further improve human-likeness, which remains our future work.

\noindent \textbf{Level-Aware Dead Code Reinitialization.}
We further introduce a level-aware reinitialization scheme that keeps both codebooks populated throughout training.
Since $\textbf{A}$ is fitted to the subaction prototypes rather than to the encoder embeddings, an unused subaction prototype does not only waste capacity at its own level: it also withdraws its contribution to the action prototypes above it, so the two levels can degrade together.
Every five epochs we therefore resample underused prototypes from the latents each level is meant to cover ($\textbf{e}_t$ for $\textbf{Z}$, $\textbf{q}^\textbf{Z}_t$ for $\textbf{A}$), setting the usage threshold per level rather than globally: $3$ for the larger codebook $\textbf{Z}$, so that rarely selected subactions are recycled, and $1$ for the smaller $\textbf{A}$, where only entirely unused prototypes are replaced.

\subsection{Reinforcement Learning with HiMAQ}

We train a policy $\pi_\theta$ to select an index from 
the subaction codebook $\textbf{Z}$, producing $\textbf{q}^\textbf{Z}_t$. The policy $\pi_\theta$ 
takes the current state $\textbf{s}_t$ as input and produces a distribution over 
$\alpha K$ logits, each corresponding to one subaction prototype in $\textbf{Z}$. 
At each state $\textbf{s}_t$, $\pi_\theta$ samples an index from this distribution, 
retrieving the corresponding $\textbf{q}^\textbf{Z}_t$. This subaction code $\textbf{q}^\textbf{Z}_t$ is then 
used to retrieve the action prototype $\textbf{q}^\textbf{A}_t$ from codebook $\textbf{A}$. With 
$\textbf{q}^\textbf{A}_t$ and $\textbf{s}_t$, the decoder reconstructs the macro action $\widetilde{\textbf{m}}_t$ 
to interact with the environment, as shown in Fig.~\ref{fig:method}(b). The policy $\pi_\theta$ is optimized to find the macro action maximizing the 
expected cumulative reward within the human manifold:
\begin{equation}
\textbf{m}^*_t = {\arg\max}_{\textbf{m}_t \in \mathcal{H}} \;
\mathbb{E}\left[\sum_{i=t}^{t+H-1} \gamma^{i-t} R(\textbf{s}_i, \textbf{a}_i) 
\;\bigg|\; \textbf{s}_t;\; \textbf{m}_t\right].
\end{equation}
Unlike MAQ~\cite{guo2026learning}, which encodes segments from $\mathcal{H}$ using 
a single-level codebook, HiMAQ encodes them through a two-level 
hierarchical quantization, yielding better action quantization/segmentation accuracy and enabling finer-grained human-like behavior.
\section{Experiments}
\label{sec:experiments}

We evaluate HiMAQ on four Adroit tasks~\cite{fu_d4rl_2021}, showing consistent human-likeness improvements over MAQ~\cite{guo2026learning} across three RL algorithms.
For fair comparisons, our experiments share the same tasks, datasets, metrics, hyperparameters, and competing methods with MAQ~\cite{guo2026learning}, on top of which HiMAQ is directly built; we refer readers to the MAQ paper~\cite{guo2026learning} for exact details on these shared components.

\noindent \textbf{Implementation Details.}
We evaluate on four Adroit tasks from D4RL~\cite{fu_d4rl_2021} using three RL algorithms: IQL~\cite{kostrikov_offline_2021}, SAC~\cite{haarnoja_soft_2018}, and RLPD~\cite{ball_efficient_2023}, each augmented with MAQ~\cite{guo2026learning} and HiMAQ to yield six human-like agents.
A behavioral cloning (BC)~\cite{atkeson_robot_1997} agent serves as an imitation-learning reference.
Of the ten agents evaluated, seven --- BC, IQL, SAC, RLPD, MAQ+IQL, MAQ+SAC, and MAQ+RLPD --- are reported using the values published by Guo et al.~\cite{guo2026learning}; the three HiMAQ variants (HiMAQ+IQL, HiMAQ+SAC, HiMAQ+RLPD) are our contributions.
We re-ran all seven baselines with the authors' released code~\cite{maq_repo} and the same hyperparameters, obtaining numbers slightly worse than the published ones; we therefore report the published values.
All agents are trained for $10^6$ steps; IQL additionally receives $10^6$ offline steps.
HiMAQ uses a hierarchical VQ-VAE with two codebooks, while MAQ uses a single-level one.
SAC variants use discrete SAC (DSAC)~\cite{zhou_revisiting_2024} to accommodate the discrete action space.
Our implementation builds directly on the codebase released by the MAQ authors~\cite{maq_repo}, which already integrates IQL, SAC, and RLPD with a single-level VQ-VAE codebook; we reuse these integrations and replace the single-level codebook with our hierarchical two-level one.
Despite the additional subaction codebook, HiMAQ's hierarchical quantization adds negligible computational cost: across all three RL algorithms, datasets, and tasks, the wall-clock training time overhead relative to MAQ remains under 3\%, adding only $\sim$5 minutes to the $\sim$3-hour SAC and RLPD training runs ($<3\%$) and $\sim$1 hour to the $\sim$2-day IQL runs ($<2\%$).

\noindent \textbf{Datasets.} 
We evaluate on four Adroit dexterous manipulation tasks from D4RL~\cite{fu_d4rl_2021}, which provide both task environments and human demonstration datasets collected via teleoperation~\cite{rajeswaran2017learning}.
Each task is controlled by a 24-DOF simulated hand and poses distinct manipulation challenges:
\textit{Door} requires the agent to grasp a handle and swing open a hinged door;
\textit{Hammer} requires the agent to pick up a hammer and drive a nail into a board;
\textit{Pen} requires the agent to reorient a pen in-hand to match a target orientation; and
\textit{Relocate} requires the agent to pick up a ball and move it to a target location.
Each task includes 25 human demonstration trajectories, which are split into training and testing sets with a 9:1 ratio.
All methods that rely on human demonstrations are trained on the same training split, and trajectory similarity is evaluated on the test split, following Guo et al.~\cite{guo2026learning}.

\noindent \textbf{Evaluation Metrics.}
We follow Guo et al.~\cite{guo2026learning} to adopt the same two trajectory similarity metrics.
\textit{Dynamic Time Warping} (DTW) measures sequential alignment between agent and human trajectories, evaluated separately on states ($\text{DTW}_s$) and actions ($\text{DTW}_a$).
\textit{Wasserstein Distance} (WD) measures distributional similarity between agent and human trajectories, likewise split into state-based ($\text{WD}_s$) and action-based ($\text{WD}_a$) variants.
For all four metrics, higher normalized scores indicate more human-like behavior.

\noindent \textbf{Reporting of Aggregate Results.}
Per-task entries report the mean and standard deviation over three random seeds.
Throughout the paper, Average rows report the mean over the four tasks together with the pooled across-seed standard deviation $s_\mathrm{pooled} = \sqrt{\tfrac{1}{T}\sum_{i=1}^{T} s_i^2}$, where $s_i$ is the seed-level standard deviation on task $i$ and $T{=}4$.
With an equal number of seeds per task this is the usual pooled estimator, and it places the aggregate on the same scale as the per-task entries.

\subsection{Quantitative Trajectory Evaluations}

\subsubsection{State-of-the-Art Comparisons}
\label{sec:performance_of_HiMAQ}

Tabs.~\ref{table:iql_results},~\ref{table:sac_results}, and~\ref{table:rlpd_results} report results across Adroit tasks under IQL~\cite{kostrikov_offline_2021}, SAC~\cite{haarnoja_soft_2018}, and RLPD~\cite{ball_efficient_2023} baselines.
Each table reports results for BC~\cite{atkeson_robot_1997}, base RL agent, and MAQ~\cite{guo2026learning} and HiMAQ variants. HiMAQ outperforms MAQ~\cite{guo2026learning} on trajectory similarity metrics, while improving success rates.
For IQL baseline (Tab.~\ref{table:iql_results}), HiMAQ+IQL improves average $\text{DTW}_s$ from 0.64 to 0.67 and success rate from 0.39 to 0.68.
Next, for SAC baseline (Tab.~\ref{table:sac_results}), HiMAQ+SAC improves average $\text{DTW}_s$ from 0.56 to 0.63 and success rate from 0.28 to 0.32.
Lastly, for RLPD baseline (Tab.~\ref{table:rlpd_results}), HiMAQ+RLPD improves average $\text{DTW}_s$ from 0.56 to 0.62 and success rate from 0.52 to 0.61.
Although the unconstrained RLPD policy achieves a higher success rate (0.68) than HiMAQ+RLPD (0.61) by not restricting actions to the human manifold, HiMAQ+RLPD (0.61) outperforms the closest competitor MAQ+RLPD (0.52) while improving trajectory similarity metrics. The above results validate the benefits of modeling subtle variations within macro actions via our hierarchical macro action quantization framework.

\begin{table}[!ht]
  \centering
  \caption{Improvements with the IQL baseline on the Adroit benchmark. Per-task rows report mean $\pm$ standard deviation across three seeds.}
   \label{table:iql_results}

 \resizebox{\columnwidth}{!}{
     \scriptsize
     \renewcommand{\arraystretch}{0.85}
  \begin{tabular}{ll||r|rrr}
  \toprule
  \textbf{Task} &  & \multicolumn{1}{c|}{\textbf{BC}} & \multicolumn{1}{c}{\textbf{IQL}} &
                       \multicolumn{1}{c}{\textbf{MAQ+IQL}}  & \multicolumn{1}{c}{\textbf{HiMAQ+IQL}} \\
\midrule    

\multirow{5}{*}{Door} & $\text{DTW}_s(\uparrow)$ & 0.18 $\pm$ 0.09 & 0.43 $\pm$ 0.06 & 0.84 $\pm$ 0.06 & \textbf{0.87  $\pm$ 0.08} \\
 & $\text{DTW}_a(\uparrow)$ & 0.42 $\pm$ 0.13 & 0.61 $\pm$ 0.04 & 0.95 $\pm$ 0.01 & \textbf{0.96 $\pm$ 0.01} \\
 & $\text{WD}_s(\uparrow)$ & 0.32 $\pm$ 0.05 & 0.48 $\pm$ 0.05 & 0.75 $\pm$ 0.05 & \textbf{0.77  $\pm$ 0.05} \\
 & $\text{WD}_a(\uparrow)$ & 0.41 $\pm$ 0.08 & 0.50 $\pm$ 0.02 & 0.81 $\pm$ 0.03 & \textbf{0.82 $\pm$ 0.03} \\
 & $\text{Success}(\uparrow)$ & 0.02 $\pm$ 0.01 & 0.16 $\pm$ 0.06 & 0.93 $\pm$ 0.04 & \textbf{0.95 $\pm$ 0.04} \\
\midrule    
\multirow{5}{*}{Hammer} & $\text{DTW}_s(\uparrow)$ & -0.16 $\pm$ 0.07 & -0.14 $\pm$ 0.34 & 0.64 $\pm$ 0.17 & \textbf{0.68 $\pm$ 0.12} \\
 & $\text{DTW}_a(\uparrow)$ & 0.47 $\pm$ 0.02 & 0.45 $\pm$ 0.20 & 0.92 $\pm$ 0.06 & \textbf{0.94 $\pm$ 0.05} \\
 & $\text{WD}_s(\uparrow)$ & 0.11 $\pm$ 0.06 & 0.12 $\pm$ 0.13 & \textbf{0.75 $\pm$ 0.03} & 0.73 $\pm$ 0.04 \\
 & $\text{WD}_a(\uparrow)$ & 0.30 $\pm$ 0.03 & 0.30 $\pm$ 0.10 & 0.84 $\pm$ 0.02 & \textbf{0.87 $\pm$ 0.02} \\
 & $\text{Success}(\uparrow)$ & 0.00 $\pm$ 0.00 & 0.01 $\pm$ 0.01 & 0.00 $\pm$ 0.00 & \textbf{0.87  $\pm$ 0.06} \\
\midrule    
\multirow{5}{*}{Pen} & $\text{DTW}_s(\uparrow)$ & 0.53 $\pm$ 0.13 & 0.34 $\pm$ 0.09 & 0.55 $\pm$ 0.17 &\textbf{0.57  $\pm$ 0.11} \\
 & $\text{DTW}_a(\uparrow)$ & 0.58 $\pm$ 0.05 & 0.51 $\pm$ 0.05 & 0.58 $\pm$ 0.09 & \textbf{0.60  $\pm$ 0.06} \\
 & $\text{WD}_s(\uparrow)$ & 0.59 $\pm$ 0.12 & 0.54 $\pm$ 0.11 & 0.59 $\pm$ 0.13 &\textbf{ 0.61  $\pm$ 0.10} \\
 & $\text{WD}_a(\uparrow)$ & 0.65 $\pm$ 0.14 & 0.63 $\pm$ 0.12 & 0.66 $\pm$ 0.15 & \textbf{0.68  $\pm$  0.11 }\\
 & $\text{Success}(\uparrow)$ & 0.40 $\pm$ 0.03 & 0.40 $\pm$ 0.05 & 0.42 $\pm$ 0.07 & \textbf{0.66  $\pm$ 0.02} \\
\midrule    
\multirow{5}{*}{Relocate} & $\text{DTW}_s(\uparrow)$ & 0.09 $\pm$ 0.14 & 0.20 $\pm$ 0.20 & 0.52 $\pm$ 0.06 & \textbf{0.54 $\pm$ 0.10} \\
 & $\text{DTW}_a(\uparrow)$ & 0.47 $\pm$ 0.15 & 0.51 $\pm$ 0.11 & 0.82 $\pm$ 0.01 & \textbf{0.84  $\pm$  0.04} \\
 & $\text{WD}_s(\uparrow)$ & 0.27 $\pm$ 0.15 & 0.36 $\pm$ 0.06 & \textbf{0.47 $\pm$ 0.07} & \textbf{0.47  $\pm$  0.05} \\
 & $\text{WD}_a(\uparrow)$ & 0.45 $\pm$ 0.12 & 0.50 $\pm$ 0.04 & 0.65 $\pm$ 0.03 & \textbf{0.66  $\pm$  0.02} \\
 & $\text{Success}(\uparrow)$ & 0.01 $\pm$ 0.02 & 0.00 $\pm$ 0.00 & 0.20 $\pm$ 0.10 & \textbf{0.23  $\pm$ 0.05}  \\
\midrule    
\midrule    
\multirow{5}{*}{Average} & $\text{DTW}_s(\uparrow)$ & \avgseed{0.18}{-0.16}{0.53}{0.09}{0.09}{0.07}{0.13}{0.14} & \avgseed{0.43}{-0.14}{0.34}{0.2}{0.06}{0.34}{0.09}{0.2} & \avgseed{0.84}{0.64}{0.55}{0.52}{0.06}{0.17}{0.17}{0.06} & \textbf{\avgseed{0.87}{0.68}{0.57}{0.54}{0.08}{0.12}{0.11}{0.1}} \\
 & $\text{DTW}_a(\uparrow)$ & \avgseed{0.42}{0.47}{0.58}{0.47}{0.13}{0.02}{0.05}{0.15} & \avgseed{0.61}{0.45}{0.51}{0.51}{0.04}{0.2}{0.05}{0.11} & \avgseed{0.95}{0.92}{0.58}{0.82}{0.01}{0.06}{0.09}{0.01} & \textbf{\avgseed{0.96}{0.94}{0.6}{0.84}{0.01}{0.05}{0.06}{0.04}} \\
 & $\text{WD}_s(\uparrow)$ & \avgseed{0.32}{0.11}{0.59}{0.27}{0.05}{0.06}{0.12}{0.15} & \avgseed{0.48}{0.12}{0.54}{0.36}{0.05}{0.13}{0.11}{0.06} & \avgseed{0.75}{0.75}{0.59}{0.47}{0.05}{0.03}{0.13}{0.07} & \textbf{\avgseed{0.77}{0.73}{0.61}{0.47}{0.05}{0.04}{0.1}{0.05}} \\
 & $\text{WD}_a(\uparrow)$ & \avgseed{0.41}{0.3}{0.65}{0.45}{0.08}{0.03}{0.14}{0.12} & \avgseed{0.5}{0.3}{0.63}{0.5}{0.02}{0.1}{0.12}{0.04} & \avgseed{0.81}{0.84}{0.66}{0.65}{0.03}{0.02}{0.15}{0.03} & \textbf{\avgseed{0.82}{0.87}{0.68}{0.66}{0.03}{0.02}{0.11}{0.02}} \\
 & $\text{Success}(\uparrow)$ & \avgseed{0.02}{0}{0.4}{0.01}{0.01}{0}{0.03}{0.02} & \avgseed{0.16}{0.01}{0.4}{0}{0.06}{0.01}{0.05}{0} & \avgseed{0.93}{0}{0.42}{0.2}{0.04}{0}{0.07}{0.1} & \textbf{\avgseed{0.95}{0.87}{0.66}{0.23}{0.04}{0.06}{0.02}{0.05}} \\
\bottomrule
  \end{tabular}
}
\end{table}
                
\begin{table}[!ht]
  \centering
  \caption{Improvements with the SAC baseline on the Adroit benchmark. Per-task rows report mean $\pm$ standard deviation across three seeds.}
   \label{table:sac_results}
 \resizebox{\columnwidth}{!}{
     \scriptsize
     \renewcommand{\arraystretch}{0.85}
  \begin{tabular}{ll||r|rrr}
  \toprule
  \textbf{Task} &  & \multicolumn{1}{c|}{\textbf{BC}} & \multicolumn{1}{c}{\textbf{SAC}} &
                       \multicolumn{1}{c}{\textbf{MAQ+SAC}}  & \multicolumn{1}{c}{\textbf{HiMAQ+SAC}} \\
\midrule    

\multirow{5}{*}{Door} 
 & $\text{DTW}_s(\uparrow)$ & 0.18 $\pm$ 0.09 & -0.39 $\pm$ 0.10 & 0.80 $\pm$ 0.08 & \textbf{0.83 $\pm$ 0.03} \\
 & $\text{DTW}_a(\uparrow)$ & 0.42 $\pm$ 0.13 & -0.25 $\pm$ 0.04 & 0.91 $\pm$ 0.03 & \textbf{0.92 $\pm$ 0.02} \\
 & $\text{WD}_s(\uparrow)$ & 0.32 $\pm$ 0.05 & -0.28 $\pm$ 0.02 & 0.71 $\pm$ 0.08 & \textbf{0.75 $\pm$ 0.03} \\
 & $\text{WD}_a(\uparrow)$ & 0.41 $\pm$ 0.08 & -0.15 $\pm$ 0.02 & 0.77 $\pm$ 0.07 & \textbf{0.80 $\pm$ 0.02} \\
 & $\text{Success}(\uparrow)$ & 0.02 $\pm$ 0.01 & 0.43 $\pm$ 0.23 & 0.56 $\pm$ 0.50 & \textbf{0.60 $\pm$ 0.29} \\
\midrule

\multirow{5}{*}{Hammer} 
 & $\text{DTW}_s(\uparrow)$ & -0.16 $\pm$ 0.07 & -1.10 $\pm$ 0.35 & 0.61 $\pm$ 0.21 & \textbf{0.64 $\pm$ 0.14} \\
 & $\text{DTW}_a(\uparrow)$ & 0.47 $\pm$ 0.02 & -0.33 $\pm$ 0.11 & 0.91 $\pm$ 0.10 & \textbf{0.92 $\pm$ 0.07} \\
 & $\text{WD}_s(\uparrow)$ & 0.11 $\pm$ 0.06 & -0.44 $\pm$ 0.09 & 0.64 $\pm$ 0.12 & \textbf{0.67 $\pm$ 0.07} \\
 & $\text{WD}_a(\uparrow)$ & 0.30 $\pm$ 0.03 & -0.19 $\pm$ 0.04 & 0.78 $\pm$ 0.11 & \textbf{0.82 $\pm$ 0.02} \\
 & $\text{Success}(\uparrow)$ & 0.00 $\pm$ 0.00 & \textbf{0.01 $\pm$ 0.01} & 0.00 $\pm$ 0.00 & 0.00 $\pm$ 0.00 \\
\midrule   

\multirow{5}{*}{Pen} 
 & $\text{DTW}_s(\uparrow)$ & 0.53 $\pm$ 0.13 & 0.06 $\pm$ 0.20 & 0.58 $\pm$ 0.17 & \textbf{0.58 $\pm$ 0.12} \\
 & $\text{DTW}_a(\uparrow)$ & 0.58 $\pm$ 0.05 & -0.34 $\pm$ 0.16 & 0.58 $\pm$ 0.11 & \textbf{0.58 $\pm$ 0.05} \\
 & $\text{WD}_s(\uparrow)$ & 0.59 $\pm$ 0.12 & 0.29 $\pm$ 0.10 & 0.61 $\pm$ 0.12 & \textbf{0.61 $\pm$ 0.10} \\
 & $\text{WD}_a(\uparrow)$ & 0.65 $\pm$ 0.14 & 0.22 $\pm$ 0.15 & 0.67 $\pm$ 0.14 & \textbf{0.67 $\pm$ 0.10} \\
 & $\text{Success}(\uparrow)$ & 0.40 $\pm$ 0.03 & 0.32 $\pm$ 0.09 & 0.41 $\pm$ 0.01 & \textbf{0.51 $\pm$ 0.04} \\
\midrule

\multirow{5}{*}{Relocate} 
 & $\text{DTW}_s(\uparrow)$ & 0.09 $\pm$ 0.14 & -0.55 $\pm$ 0.20 & 0.25 $\pm$ 0.19 & \textbf{0.46 $\pm$ 0.07} \\
 & $\text{DTW}_a(\uparrow)$ & 0.47 $\pm$ 0.15 & -0.10 $\pm$ 0.16 & 0.66 $\pm$ 0.09 & \textbf{0.80 $\pm$ 0.05} \\
 & $\text{WD}_s(\uparrow)$ & 0.27 $\pm$ 0.15 & -0.22 $\pm$ 0.06 & 0.40 $\pm$ 0.07 & \textbf{0.47 $\pm$ 0.08} \\
 & $\text{WD}_a(\uparrow)$ & 0.45 $\pm$ 0.12 & -0.05 $\pm$ 0.03 & 0.61 $\pm$ 0.03 & \textbf{0.66 $\pm$ 0.05} \\
 & $\text{Success}(\uparrow)$ & 0.01 $\pm$ 0.02 & 0.00 $\pm$ 0.00 & 0.14 $\pm$ 0.07 & \textbf{0.16 $\pm$ 0.08} \\
\midrule  

\midrule    

\multirow{5}{*}{Average} 
 & $\text{DTW}_s(\uparrow)$ & \avgseed{0.18}{-0.16}{0.53}{0.09}{0.09}{0.07}{0.13}{0.14} & \avgseed{-0.39}{-1.1}{0.06}{-0.55}{0.1}{0.35}{0.2}{0.2} & \avgseed{0.8}{0.61}{0.58}{0.25}{0.08}{0.21}{0.17}{0.19} & \textbf{\avgseed{0.83}{0.64}{0.58}{0.46}{0.03}{0.14}{0.12}{0.07}} \\
 & $\text{DTW}_a(\uparrow)$ & \avgseed{0.42}{0.47}{0.58}{0.47}{0.13}{0.02}{0.05}{0.15} & \avgseed{-0.25}{-0.33}{-0.34}{-0.1}{0.04}{0.11}{0.16}{0.16} & \avgseed{0.91}{0.91}{0.58}{0.66}{0.03}{0.1}{0.11}{0.09} & \textbf{\avgseed{0.92}{0.92}{0.58}{0.8}{0.02}{0.07}{0.05}{0.05}} \\
 & $\text{WD}_s(\uparrow)$ & \avgseed{0.32}{0.11}{0.59}{0.27}{0.05}{0.06}{0.12}{0.15} & \avgseed{-0.28}{-0.44}{0.29}{-0.22}{0.02}{0.09}{0.1}{0.06} & \avgseed{0.71}{0.64}{0.61}{0.4}{0.08}{0.12}{0.12}{0.07} & \textbf{\avgseed{0.75}{0.67}{0.61}{0.47}{0.03}{0.07}{0.1}{0.08}} \\
 & $\text{WD}_a(\uparrow)$ & \avgseed{0.41}{0.3}{0.65}{0.45}{0.08}{0.03}{0.14}{0.12} & \avgseed{-0.15}{-0.19}{0.22}{-0.05}{0.02}{0.04}{0.15}{0.03} & \avgseed{0.77}{0.78}{0.67}{0.61}{0.07}{0.11}{0.14}{0.03} & \textbf{\avgseed{0.8}{0.82}{0.67}{0.66}{0.02}{0.02}{0.1}{0.05}} \\
 & $\text{Success}(\uparrow)$ & \avgseed{0.02}{0}{0.4}{0.01}{0.01}{0}{0.03}{0.02} & \avgseed{0.43}{0.01}{0.32}{0}{0.23}{0.01}{0.09}{0} & \avgseed{0.56}{0}{0.41}{0.14}{0.5}{0}{0.01}{0.07} & \textbf{\avgseed{0.6}{0}{0.51}{0.16}{0.29}{0}{0.04}{0.08}} \\
\bottomrule
  \end{tabular}
}
\end{table}
\begin{table}[!ht]
  \centering
  \caption{Improvements with the RLPD baseline on the Adroit benchmark. Per-task rows report mean $\pm$ standard deviation across three seeds.}
   \label{table:rlpd_results}
 \resizebox{\columnwidth}{!}{
     \scriptsize
     \renewcommand{\arraystretch}{0.85}
  \begin{tabular}{ll||r|rrr}
  \toprule
  \textbf{Task} &  & \multicolumn{1}{c|}{\textbf{BC}} & \multicolumn{1}{c}{\textbf{RLPD}} &
                       \multicolumn{1}{c}{\textbf{MAQ+RLPD}}  & \multicolumn{1}{c}{\textbf{HiMAQ+RLPD}} \\
\midrule    

\multirow{5}{*}{Door} 
 & $\text{DTW}_s(\uparrow)$ & 0.18 $\pm$ 0.09 & -0.06 $\pm$ 0.04 & 0.76 $\pm$ 0.04 & \textbf{0.83 $\pm$ 0.07} \\
 & $\text{DTW}_a(\uparrow)$ & 0.42 $\pm$ 0.13 & 0.28 $\pm$ 0.08 & 0.91 $\pm$ 0.05 & \textbf{0.93 $\pm$ 0.01} \\
 & $\text{WD}_s(\uparrow)$ & 0.32 $\pm$ 0.05 & -0.14 $\pm$ 0.04 & 0.71 $\pm$ 0.03 & \textbf{0.76 $\pm$ 0.04} \\
 & $\text{WD}_a(\uparrow)$ & 0.41 $\pm$ 0.08 & 0.10 $\pm$ 0.02 & 0.76 $\pm$ 0.03 & \textbf{0.79 $\pm$ 0.03} \\
 & $\text{Success}(\uparrow)$ & 0.02 $\pm$ 0.01 & \textbf{0.96 $\pm$ 0.07} & 0.93 $\pm$ 0.05 & \textbf{0.96 $\pm$ 0.10} \\
\midrule    

\multirow{5}{*}{Hammer} 
 & $\text{DTW}_s(\uparrow)$ & -0.16 $\pm$ 0.07 & -0.03 $\pm$ 0.14 & \textbf{0.68 $\pm$ 0.17} & 0.67 $\pm$ 0.12 \\
 & $\text{DTW}_a(\uparrow)$ & 0.47 $\pm$ 0.02 & 0.37 $\pm$ 0.08 & 0.94 $\pm$ 0.07 & \textbf{0.94 $\pm$ 0.05} \\
 & $\text{WD}_s(\uparrow)$ & 0.11 $\pm$ 0.06 & -0.03 $\pm$ 0.08 & 0.76 $\pm$ 0.04 & \textbf{0.79 $\pm$ 0.04} \\
 & $\text{WD}_a(\uparrow)$ & 0.30 $\pm$ 0.03 & 0.20 $\pm$ 0.03 & 0.85 $\pm$ 0.03 & \textbf{0.88 $\pm$ 0.04} \\
 & $\text{Success}(\uparrow)$ & 0.00 $\pm$ 0.00 & \textbf{1.00 $\pm$ 0.00} & 0.56 $\pm$ 0.37 & 0.62 $\pm$ 0.03 \\
\midrule    

\multirow{5}{*}{Pen} 
 & $\text{DTW}_s(\uparrow)$ & 0.53 $\pm$ 0.13 & 0.48 $\pm$ 0.24 & 0.54 $\pm$ 0.18 & \textbf{0.61 $\pm$ 0.11} \\
 & $\text{DTW}_a(\uparrow)$ & 0.58 $\pm$ 0.05 & 0.40 $\pm$ 0.17 & 0.59 $\pm$ 0.13 & \textbf{0.64 $\pm$ 0.08} \\
 & $\text{WD}_s(\uparrow)$ & 0.59 $\pm$ 0.12 & 0.49 $\pm$ 0.08 & 0.59 $\pm$ 0.12 & \textbf{0.62 $\pm$ 0.06} \\
 & $\text{WD}_a(\uparrow)$ & 0.65 $\pm$ 0.14 & 0.44 $\pm$ 0.12 & 0.66 $\pm$ 0.14 & \textbf{0.69 $\pm$ 0.08} \\
 & $\text{Success}(\uparrow)$ & 0.40 $\pm$ 0.03 & 0.62 $\pm$ 0.09 & 0.42 $\pm$ 0.05 & \textbf{0.64 $\pm$ 0.06} \\
\midrule    

\multirow{5}{*}{Relocate} 
 & $\text{DTW}_s(\uparrow)$ & 0.09 $\pm$ 0.14 & 0.03 $\pm$ 0.13 & 0.27 $\pm$ 0.14 & \textbf{0.37 $\pm$ 0.07} \\
 & $\text{DTW}_a(\uparrow)$ & 0.47 $\pm$ 0.15 & 0.32 $\pm$ 0.13 & 0.69 $\pm$ 0.10 & \textbf{0.74 $\pm$ 0.01} \\
 & $\text{WD}_s(\uparrow)$ & 0.27 $\pm$ 0.15 & 0.02 $\pm$ 0.07 & 0.38 $\pm$ 0.09 & \textbf{0.42 $\pm$ 0.07} \\
 & $\text{WD}_a(\uparrow)$ & 0.45 $\pm$ 0.12 & 0.20 $\pm$ 0.03 & 0.55 $\pm$ 0.08 & \textbf{0.60 $\pm$ 0.05} \\
 & $\text{Success}(\uparrow)$ & 0.01 $\pm$ 0.02 & 0.14 $\pm$ 0.03 & 0.17 $\pm$ 0.10 & \textbf{0.23 $\pm$ 0.05}  \\
\midrule    

\midrule    

\multirow{5}{*}{Average} 
 & $\text{DTW}_s(\uparrow)$ & \avgseed{0.18}{-0.16}{0.53}{0.09}{0.09}{0.07}{0.13}{0.14} & \avgseed{-0.06}{-0.03}{0.48}{0.03}{0.04}{0.14}{0.24}{0.13} & \avgseed{0.76}{0.68}{0.54}{0.27}{0.04}{0.17}{0.18}{0.14} & \textbf{\avgseed{0.83}{0.67}{0.61}{0.37}{0.07}{0.12}{0.11}{0.07}} \\
 & $\text{DTW}_a(\uparrow)$ & \avgseed{0.42}{0.47}{0.58}{0.47}{0.13}{0.02}{0.05}{0.15} & \avgseed{0.28}{0.37}{0.4}{0.32}{0.08}{0.08}{0.17}{0.13} & \avgseed{0.91}{0.94}{0.59}{0.69}{0.05}{0.07}{0.13}{0.1} & \textbf{\avgseed{0.93}{0.94}{0.64}{0.74}{0.01}{0.05}{0.08}{0.01}} \\
 & $\text{WD}_s(\uparrow)$ & \avgseed{0.32}{0.11}{0.59}{0.27}{0.05}{0.06}{0.12}{0.15} & \avgseed{-0.14}{-0.03}{0.49}{0.02}{0.04}{0.08}{0.08}{0.07} & \avgseed{0.71}{0.76}{0.59}{0.38}{0.03}{0.04}{0.12}{0.09} & \textbf{\avgseed{0.76}{0.79}{0.62}{0.42}{0.04}{0.04}{0.06}{0.07}} \\
 & $\text{WD}_a(\uparrow)$ & \avgseed{0.41}{0.3}{0.65}{0.45}{0.08}{0.03}{0.14}{0.12} & \avgseed{0.1}{0.2}{0.44}{0.2}{0.02}{0.03}{0.12}{0.03} & \avgseed{0.76}{0.85}{0.66}{0.55}{0.03}{0.03}{0.14}{0.08} & \textbf{\avgseed{0.79}{0.88}{0.69}{0.6}{0.03}{0.04}{0.08}{0.05}} \\
 & $\text{Success}(\uparrow)$ & \avgseed{0.02}{0}{0.4}{0.01}{0.01}{0}{0.03}{0.02} & \textbf{\avgseed{0.96}{1}{0.62}{0.14}{0.07}{0}{0.09}{0.03}} & \avgseed{0.93}{0.56}{0.42}{0.17}{0.05}{0.37}{0.05}{0.1} & \avgseed{0.96}{0.62}{0.64}{0.23}{0.1}{0.03}{0.06}{0.05} \\
\bottomrule
  \end{tabular}
}
\end{table}

\subsubsection{Impacts of Action Lengths}
\label{sec:ablation}
\begin{figure}[!ht]
  \centering
    \includegraphics[width=\columnwidth, trim = 0mm 0mm 0mm 0mm, clip]{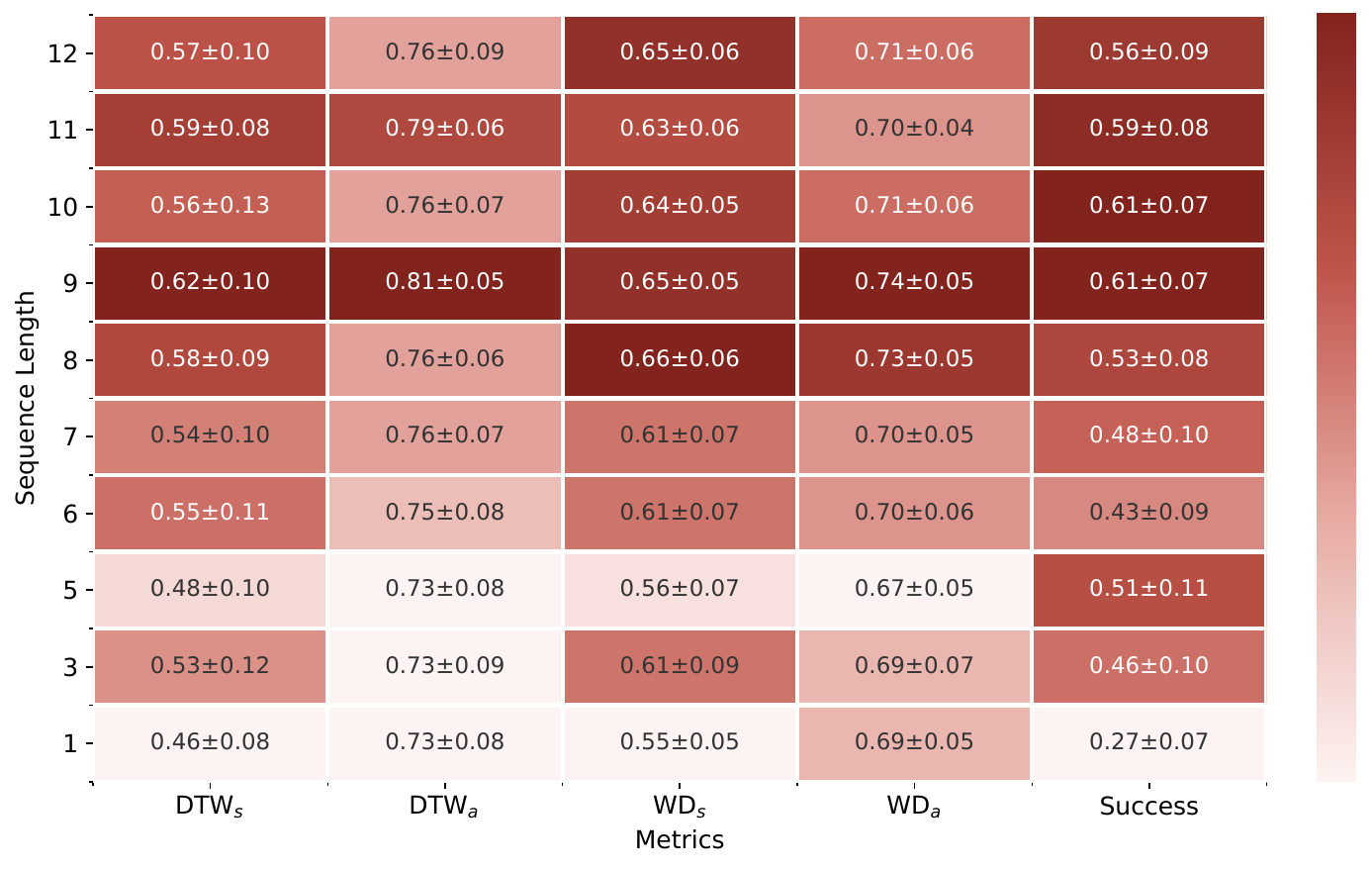}
  \caption{Impacts of different macro action lengths $H$ in HiMAQ+RLPD. Each cell shows the mean over the four Adroit tasks; color is normalized within each column.}
  \label{fig:RLPD_sequence_ablation_study}
\end{figure}

Fig.~\ref{fig:RLPD_sequence_ablation_study} shows how macro action length $H$ affects HiMAQ+RLPD performance, averaged over four Adroit tasks.
$H{=}9$ achieves the best overall balance, with the highest success rate (0.61) and strong similarity scores ($\text{DTW}_s{=}0.62$, $\text{DTW}_a{=}0.81$, $\text{WD}_s{=}0.65$, $\text{WD}_a{=}0.74$).
This optimal value of $H{=}9$ for RLPD matches the value used by MAQ+RLPD~\cite{guo2026learning}; more generally, HiMAQ inherits MAQ's per-algorithm optimal macro action length ($H{=}9$ for IQL and RLPD, $H{=}8$ for SAC), confirming that hierarchical extension does not alter the optimal macro action length.
Performance improves from $H{=}1$ to $9$, then plateaus or declines --- success drops to 0.56 at $H{=}12$ --- indicating diminishing returns for excessively long chunks.

\subsubsection{Impact of Subaction Codebook Ratio $\alpha$}
\label{sec:ablation_alpha}
\begin{figure*}[t]
  \centering
  \begin{subfigure}[b]{0.3\textwidth}
    \includegraphics[width=\linewidth]{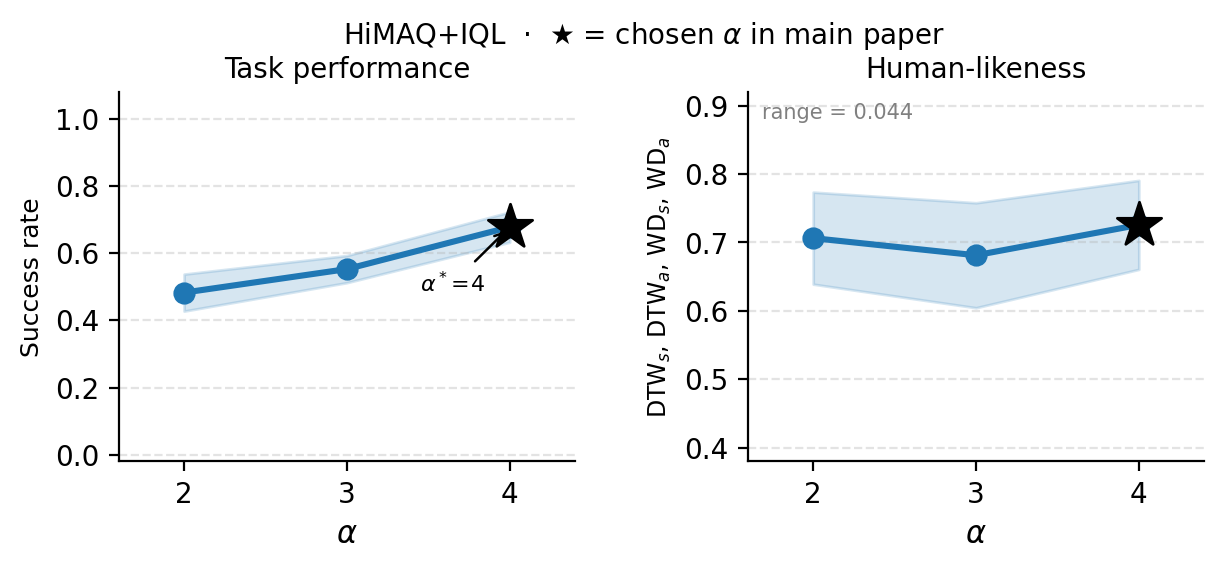}
    \caption{HiMAQ+IQL}
    \label{fig:alpha_iql}
  \end{subfigure}
  \hfill
  \begin{subfigure}[b]{0.3\textwidth}
    \includegraphics[width=\linewidth]{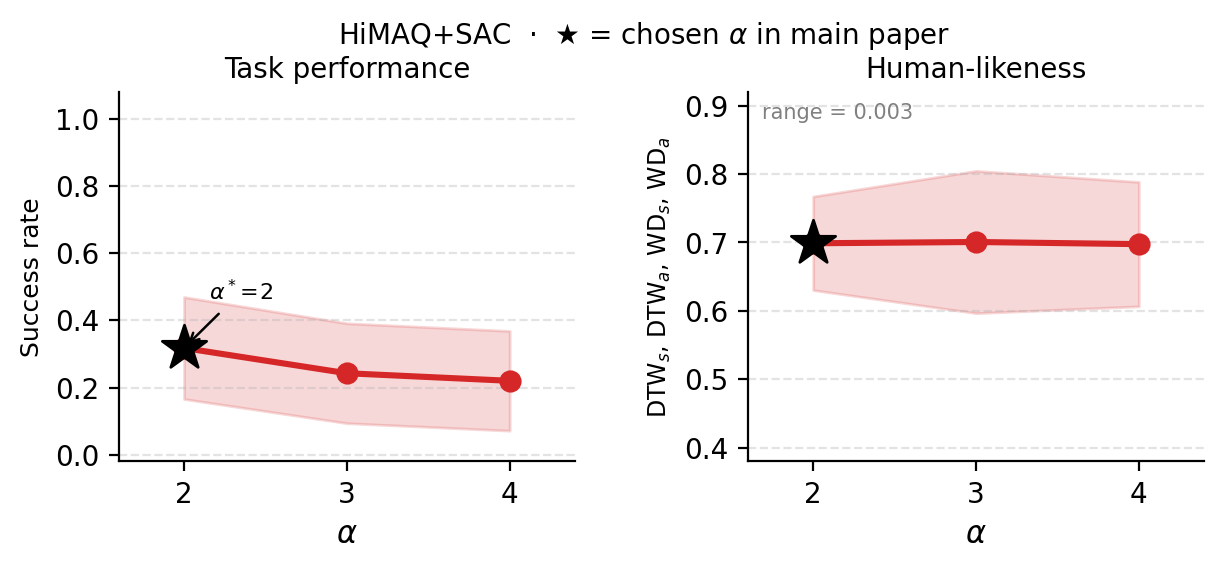}
    \caption{HiMAQ+SAC}
    \label{fig:alpha_sac}
  \end{subfigure}
  \hfill
  \begin{subfigure}[b]{0.3\textwidth}
    \includegraphics[width=\linewidth]{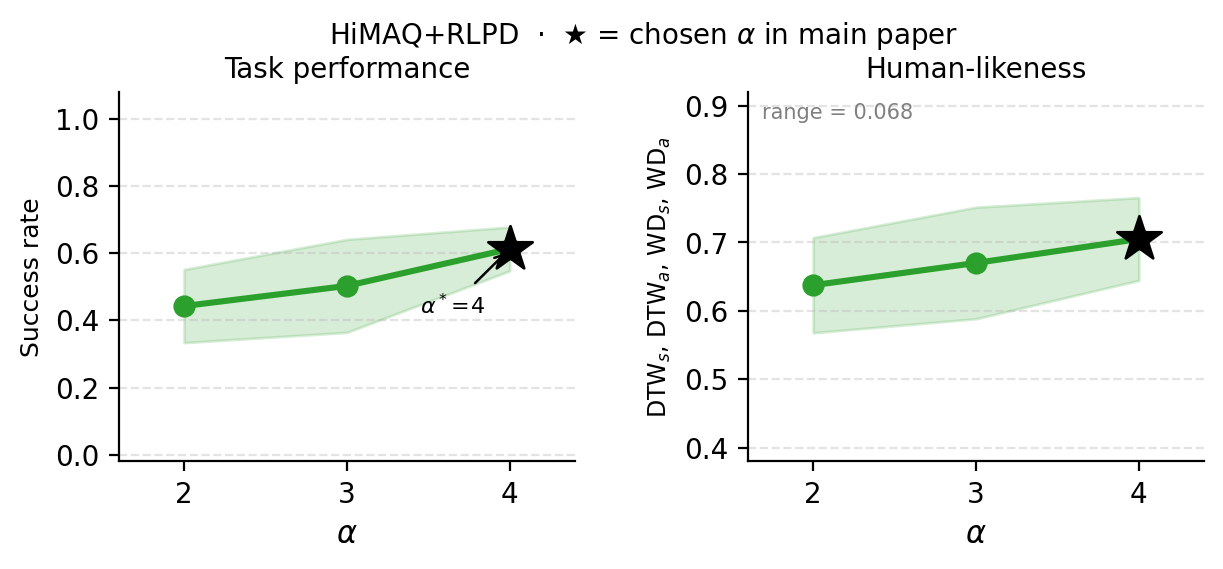}
    \caption{HiMAQ+RLPD}
    \label{fig:alpha_rlpd}
  \end{subfigure}
  \caption{Impact of the subaction codebook ratio $\alpha$ on task performance (Success rate averaged over four Adroit tasks) and mean human-likeness (average of $\text{DTW}_s$, $\text{DTW}_a$, $\text{WD}_s$, $\text{WD}_a$). Shaded bands denote $\pm 1$ standard deviation; $\bigstar$ marks the $\alpha$ used in our main results.}
  \label{fig:alpha_ablation}
\end{figure*}

The ratio $\alpha$ controls the size of the subaction codebook $\textbf{Z}$ ($\alpha K$ codes) relative to the action codebook $\textbf{A}$ ($K$ codes): a larger $\alpha$ gives the policy finer-grained action resolution at the cost of a larger discrete action space.
We sweep $\alpha \in \{2,3,4\}$ for each RL algorithm --- IQL and RLPD use $K{=}16$ (so $\alpha K \in \{32,48,64\}$), while SAC uses $K{=}8$ (so $\alpha K \in \{16,24,32\}$) --- and find that the optimal value is not universal but depends on the algorithm's capacity to exploit a richer subaction space.
Algorithms with access to offline demonstrations throughout training, IQL and RLPD, benefit from a larger codebook ($\alpha{=}4$); e.g., HiMAQ+IQL's success rate on \textit{Hammer} rises from 0.27 to 0.87 as $\alpha$ increases from 2 to 4.
The purely online HiMAQ+SAC, in contrast, is better matched to a smaller codebook ($\alpha{=}2$), as its limited exploration capacity is easily overwhelmed by a larger action space.
Notably, human-likeness metrics vary by less than 0.05 across $\alpha$ values while success rates vary substantially, indicating $\alpha$ chiefly governs performance; we accordingly set $\alpha{=}4$ for HiMAQ+IQL and HiMAQ+RLPD, and $\alpha{=}2$ for HiMAQ+SAC, for all results reported above.
Tab.~\ref{table:alpha_ablation} and Fig.~\ref{fig:alpha_ablation} summarize these trends across all three algorithms.

\begin{table}[!ht]
  \centering
  \caption{Impact of the subaction codebook ratio $\alpha$, averaged across the four Adroit tasks. We found $\alpha{=}4$ to be best for HiMAQ+IQL and HiMAQ+RLPD, and $\alpha{=}2$ to be best for HiMAQ+SAC.}
  \label{table:alpha_ablation}

  \resizebox{0.9\columnwidth}{!}{
    \scriptsize
    \renewcommand{\arraystretch}{0.85}
  \begin{tabular}{llccc}
  \toprule
  \textbf{Algorithm} & \textbf{Metric} & $\alpha=2$ & $\alpha=3$ & $\alpha=4$ \\
\midrule
\multirow{5}{*}{HiMAQ+IQL}
  & $\text{DTW}_s(\uparrow)$   & $\avgseedm{0.85}{0.63}{0.54}{0.46}{0.04}{0.09}{0.1}{0.14}$ & $\avgseedm{0.86}{0.58}{0.56}{0.38}{0.07}{0.08}{0.1}{0.18}$ & \textbf{$\avgseedm{0.87}{0.68}{0.57}{0.54}{0.08}{0.12}{0.11}{0.1}$} \\
  & $\text{DTW}_a(\uparrow)$   & $\avgseedm{0.95}{0.94}{0.58}{0.79}{0.02}{0.05}{0.07}{0.09}$ & $\avgseedm{0.96}{0.92}{0.6}{0.76}{0.01}{0.08}{0.07}{0.09}$ & \textbf{$\avgseedm{0.96}{0.94}{0.6}{0.84}{0.01}{0.05}{0.06}{0.04}$} \\
  & $\text{WD}_s(\uparrow)$    & \textbf{$\avgseedm{0.78}{0.75}{0.61}{0.46}{0.03}{0.06}{0.09}{0.06}$} & $\avgseedm{0.78}{0.59}{0.62}{0.44}{0.04}{0.1}{0.1}{0.04}$ & \textbf{$\avgseedm{0.77}{0.73}{0.61}{0.47}{0.05}{0.04}{0.1}{0.05}$} \\
  & $\text{WD}_a(\uparrow)$    & $\avgseedm{0.81}{0.85}{0.67}{0.63}{0.02}{0.01}{0.11}{0.02}$ & $\avgseedm{0.81}{0.76}{0.69}{0.59}{0.04}{0.05}{0.12}{0.01}$ & \textbf{$\avgseedm{0.82}{0.87}{0.68}{0.66}{0.03}{0.02}{0.11}{0.02}$} \\
  & $\text{Success}(\uparrow)$ & $\avgseedm{0.92}{0.27}{0.62}{0.12}{0.06}{0}{0.09}{0.02}$ & $\avgseedm{0.95}{0.5}{0.7}{0.06}{0.03}{0.02}{0.04}{0.06}$ & \textbf{$\avgseedm{0.95}{0.87}{0.66}{0.23}{0.04}{0.06}{0.02}{0.05}$} \\
\midrule
\multirow{5}{*}{HiMAQ+SAC}
  & $\text{DTW}_s(\uparrow)$   & $\avgseedm{0.83}{0.64}{0.58}{0.46}{0.03}{0.14}{0.12}{0.07}$ & \textbf{$\avgseedm{0.78}{0.8}{0.57}{0.44}{0.02}{0.17}{0.17}{0.2}$} & $\avgseedm{0.69}{0.88}{0.59}{0.39}{0.06}{0.07}{0.18}{0.26}$ \\
  & $\text{DTW}_a(\uparrow)$   & \textbf{$\avgseedm{0.92}{0.92}{0.58}{0.8}{0.02}{0.07}{0.05}{0.05}$} & $\avgseedm{0.89}{0.92}{0.57}{0.81}{0.04}{0.09}{0.11}{0.13}$ & $\avgseedm{0.84}{0.97}{0.56}{0.74}{0.04}{0.03}{0.17}{0.16}$ \\
  & $\text{WD}_s(\uparrow)$    & \textbf{$\avgseedm{0.75}{0.67}{0.61}{0.47}{0.03}{0.07}{0.1}{0.08}$} & $\avgseedm{0.66}{0.79}{0.6}{0.45}{0.02}{0.14}{0.09}{0.06}$ & $\avgseedm{0.62}{0.86}{0.61}{0.43}{0.04}{0.1}{0.06}{0.02}$ \\
  & $\text{WD}_a(\uparrow)$    & \textbf{$\avgseedm{0.8}{0.82}{0.67}{0.66}{0.02}{0.02}{0.1}{0.05}$} & $\avgseedm{0.78}{0.86}{0.67}{0.62}{0.01}{0.11}{0.09}{0.07}$ & $\avgseedm{0.70}{0.83}{0.66}{0.65}{0.03}{0.04}{0.07}{0.02}$ \\
  & $\text{Success}(\uparrow)$ & \textbf{$\avgseedm{0.6}{0}{0.51}{0.16}{0.29}{0}{0.04}{0.08}$} & $\avgseedm{0.35}{0}{0.51}{0.11}{0.27}{0.05}{0.07}{0.09}$ & $\avgseedm{0.18}{0.07}{0.54}{0.09}{0.26}{0}{0.06}{0.13}$ \\
\midrule
\multirow{5}{*}{HiMAQ+RLPD}
  & $\text{DTW}_s(\uparrow)$   & $\avgseedm{0.76}{0.62}{0.59}{0.14}{0.06}{0.03}{0.14}{0.08}$ & $\avgseedm{0.78}{0.67}{0.61}{0.28}{0.07}{0.12}{0.12}{0.14}$ & \textbf{$\avgseedm{0.83}{0.67}{0.61}{0.37}{0.07}{0.12}{0.11}{0.07}$} \\
  & $\text{DTW}_a(\uparrow)$   & $\avgseedm{0.91}{0.89}{0.61}{0.55}{0.03}{0.02}{0.11}{0.06}$ & $\avgseedm{0.91}{0.88}{0.63}{0.67}{0.01}{0.06}{0.07}{0.07}$ & \textbf{$\avgseedm{0.93}{0.94}{0.64}{0.74}{0.01}{0.05}{0.08}{0.01}$} \\
  & $\text{WD}_s(\uparrow)$    & $\avgseedm{0.71}{0.71}{0.62}{0.35}{0.07}{0.05}{0.08}{0.07}$ & $\avgseedm{0.74}{0.77}{0.6}{0.37}{0.07}{0.11}{0.1}{0.09}$ & \textbf{$\avgseedm{0.76}{0.79}{0.62}{0.42}{0.04}{0.04}{0.06}{0.07}$} \\
  & $\text{WD}_a(\uparrow)$    & $\avgseedm{0.76}{0.81}{0.68}{0.49}{0.04}{0.04}{0.09}{0.06}$ & $\avgseedm{0.8}{0.83}{0.66}{0.52}{0.02}{0.06}{0.08}{0.07}$ & \textbf{$\avgseedm{0.79}{0.88}{0.69}{0.6}{0.03}{0.04}{0.08}{0.05}$} \\
  & $\text{Success}(\uparrow)$ & $\avgseedm{0.86}{0.31}{0.48}{0.12}{0.09}{0.16}{0.11}{0.04}$ & $\avgseedm{0.88}{0.44}{0.54}{0.15}{0.09}{0.24}{0.1}{0.03}$ & \textbf{$\avgseedm{0.96}{0.62}{0.64}{0.23}{0.1}{0.03}{0.06}{0.05}$} \\
\bottomrule
  \end{tabular}
  }
\end{table}

\subsubsection{Impact of Hierarchical Action Prediction}
\label{sec:ablation_qtz}
In the RL stage, $\pi_\theta$ predicts the subaction code $\textbf{q}^\textbf{Z}_t$ and derives $\textbf{q}^\textbf{A}_t$ as its nearest action prototype, as adopted throughout this paper; an alternative is to have $\pi_\theta$ predict $\textbf{q}^\textbf{A}_t$ directly, bypassing the subaction code, as in MAQ~\cite{guo2026learning}.
Both variants share the same pre-trained codebooks and decoder, differing only in this prediction stage.
As shown in Tab.~\ref{table:ablation_decoding}, on HiMAQ+RLPD, coarse-to-fine prediction ($\textbf{q}^\textbf{Z}_t \rightarrow \textbf{q}^\textbf{A}_t$) outperforms direct prediction of $\textbf{q}^\textbf{A}_t$ on all five metrics, improving average $\text{DTW}_s$ ($0.58 \to 0.62$), $\text{DTW}_a$ ($0.78 \to 0.81$), $\text{WD}_s$ ($0.61 \to 0.65$), $\text{WD}_a$ ($0.71 \to 0.74$), and success rate ($0.51 \to 0.61$). This confirms the advantage of our coarse-to-fine prediction.

\begin{table}[!ht]
\centering
\caption{Ablation on the RL-stage prediction for HiMAQ+RLPD: predicting subaction code first $\mathbf{q}^\mathbf{Z}_t \rightarrow \mathbf{q}^\mathbf{A}_t$ vs. predicting action code $\mathbf{q}^\mathbf{A}_t$ directly.}
\label{table:ablation_decoding}
\resizebox{0.65\columnwidth}{!}{
\tiny
\renewcommand{\arraystretch}{0.85}
\begin{tabular}{lcc}
\toprule
\textbf{Metric} & $\mathbf{q}^\mathbf{Z}_t \rightarrow \mathbf{q}^\mathbf{A}_t$ & $\mathbf{q}^\mathbf{A}_t$ \\
\midrule
$\text{DTW}_s(\uparrow)$ & \textbf{$\avgseedm{0.83}{0.67}{0.61}{0.37}{0.07}{0.12}{0.11}{0.07}$} & $\avgseedm{0.8}{0.63}{0.57}{0.32}{0.06}{0.1}{0.14}{0.09}$ \\
$\text{DTW}_a(\uparrow)$ & \textbf{$\avgseedm{0.93}{0.94}{0.64}{0.74}{0.01}{0.05}{0.08}{0.01}$} & $\avgseedm{0.92}{0.9}{0.6}{0.69}{0.02}{0.06}{0.1}{0.04}$ \\
$\text{WD}_s(\uparrow)$  & \textbf{$\avgseedm{0.76}{0.79}{0.62}{0.42}{0.04}{0.04}{0.06}{0.07}$} & $\avgseedm{0.73}{0.74}{0.6}{0.38}{0.05}{0.06}{0.07}{0.1}$ \\
$\text{WD}_a(\uparrow)$  & \textbf{$\avgseedm{0.79}{0.88}{0.69}{0.6}{0.03}{0.04}{0.08}{0.05}$} & $\avgseedm{0.78}{0.85}{0.65}{0.55}{0.03}{0.03}{0.09}{0.07}$ \\
$\text{Success}(\uparrow)$ & \textbf{$\avgseedm{0.96}{0.62}{0.64}{0.23}{0.1}{0.03}{0.06}{0.05}$} & $\avgseedm{0.9}{0.5}{0.55}{0.1}{0.08}{0.06}{0.08}{0.06}$ \\
\bottomrule
\end{tabular}
}
\end{table}

\subsubsection{Impact of Decoder Input}
\label{sec:ablation_decoder_input}
We study the impact of different input representations to the decoder, i.e., subaction prototype $\textbf{q}^\textbf{Z}_t$, action prototype $\textbf{q}^\textbf{A}_t$, or both. Each variant requires retraining both the VQ-VAE and the RL policy. 
Tab.~\ref{table:ablation_decoder_input} reports the results on HiMAQ+RLPD. It can be seen from the results that the decoder performs best when using only the output of the second vector quantization, $\textbf{q}^\textbf{A}_t$, as input. Combining $\textbf{q}^\textbf{Z}_t$ and $\textbf{q}^\textbf{A}_t$ slightly reduces performance, whereas using $\textbf{q}^\textbf{Z}_t$ alone leads to a substantial decline.

\begin{table}[!ht]
  \centering
  \caption{Ablation on the decoder input for HiMAQ+RLPD: decoding from subaction prototype $\mathbf{q}^\mathbf{Z}_t$, action prototype $\mathbf{q}^\mathbf{A}_t$, or both. Entries report the mean over the four Adroit tasks with the pooled across-seed standard deviation.}
  \label{table:ablation_decoder_input}

  \resizebox{0.9\columnwidth}{!}{
    \tiny
    \renewcommand{\arraystretch}{0.85}
  \begin{tabular}{lccc}
  \toprule
  \textbf{Metric} & $\mathbf{q}^\mathbf{Z}_t$ & $\mathbf{q}^\mathbf{A}_t$ & Both \\
\midrule
$\text{DTW}_s(\uparrow)$   & $0.53 \pm 0.10$ & \textbf{$\avgseedm{0.83}{0.67}{0.61}{0.37}{0.07}{0.12}{0.11}{0.07}$} & $0.60 \pm 0.09$ \\
$\text{DTW}_a(\uparrow)$   & $0.70 \pm 0.06$ & \textbf{$\avgseedm{0.93}{0.94}{0.64}{0.74}{0.01}{0.05}{0.08}{0.01}$} & $0.78 \pm 0.06$ \\
$\text{WD}_s(\uparrow)$    & $0.58 \pm 0.06$ & \textbf{$\avgseedm{0.76}{0.79}{0.62}{0.42}{0.04}{0.04}{0.06}{0.07}$} & $0.63 \pm 0.06$ \\
$\text{WD}_a(\uparrow)$    & $0.66 \pm 0.06$ & $\avgseedm{0.79}{0.88}{0.69}{0.6}{0.03}{0.04}{0.08}{0.05}$ & $\mathbf{0.75 \pm 0.05}$ \\
$\text{Success}(\uparrow)$ & $0.52 \pm 0.08$ & \textbf{$\avgseedm{0.96}{0.62}{0.64}{0.23}{0.1}{0.03}{0.06}{0.05}$} & $0.60 \pm 0.07$ \\
\bottomrule
  \end{tabular}
  }
\end{table}

\subsection{Human Evaluations}
\label{sec:human_evaluation_study}
To verify if trajectory-level alignment is perceptually recognized as human-like, we conducted human evaluations with 22 evaluators across Turing test and human-likeness ranking test.
Our survey setup closely follows the human evaluation protocol of MAQ~\cite{guo2026learning}: for each agent-task pair, we render all 25 trajectories, rank them by a score combining task success and human-likeness, and sample the video clip for each trial uniformly from the resulting top-5 pool, applying this identically to every agent so that no agent is favored at selection time.
We refer readers to MAQ~\cite{guo2026learning} for the full survey interface and additional setup details.

\begin{figure*}[!ht]
    \centering
    \includegraphics[width=0.9\linewidth, trim = 0mm 0mm 0mm 0mm, clip]{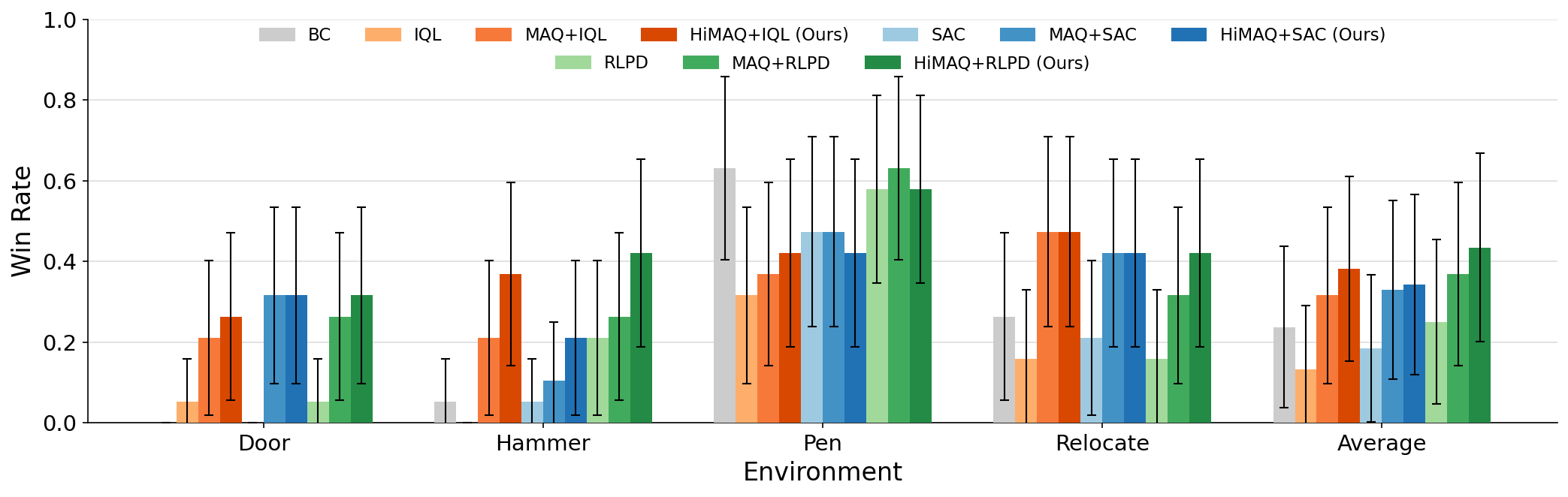}
    \caption{Turing test win rates (i.e., fraction of trials where evaluators mistook an agent for a human).}
    \label{fig:human_survey_phase_1_result}
\end{figure*}

\subsubsection{Turing Tests}
Turing test measures whether human evaluators can distinguish agent behavior from real human demonstrations.
Each evaluator is shown 24 trials, each presenting two side-by-side video clips --- one from a human demonstrator and one from a trained agent --- and asked to identify which is human.
A higher \textit{win rate} (i.e., the fraction of trials in which evaluators are fooled) indicates more human-like behavior.
This test directly probes perceptual indistinguishability, complementing the quantitative trajectory metrics. Fig.~\ref{fig:human_survey_phase_1_result} shows the Turing Test win rates.
MAQ~\cite{guo2026learning} and HiMAQ-based agents substantially outperform base RL agents, with HiMAQ variants topping the ranking:
HiMAQ+RLPD (43\%) $>$ HiMAQ+IQL (38\%) $>$ MAQ+RLPD (37\%) $>$ HiMAQ+SAC (34\%)  $>$ MAQ+SAC (33\%) $>$ MAQ+IQL (32\%) $>$ RLPD (25\%) $>$ BC (24\%) $>$ SAC (18\%) $>$ IQL (13\%).

\subsubsection{Ranking Tests}
\begin{figure}[!ht]
    \centering
    \includegraphics[width=\columnwidth]{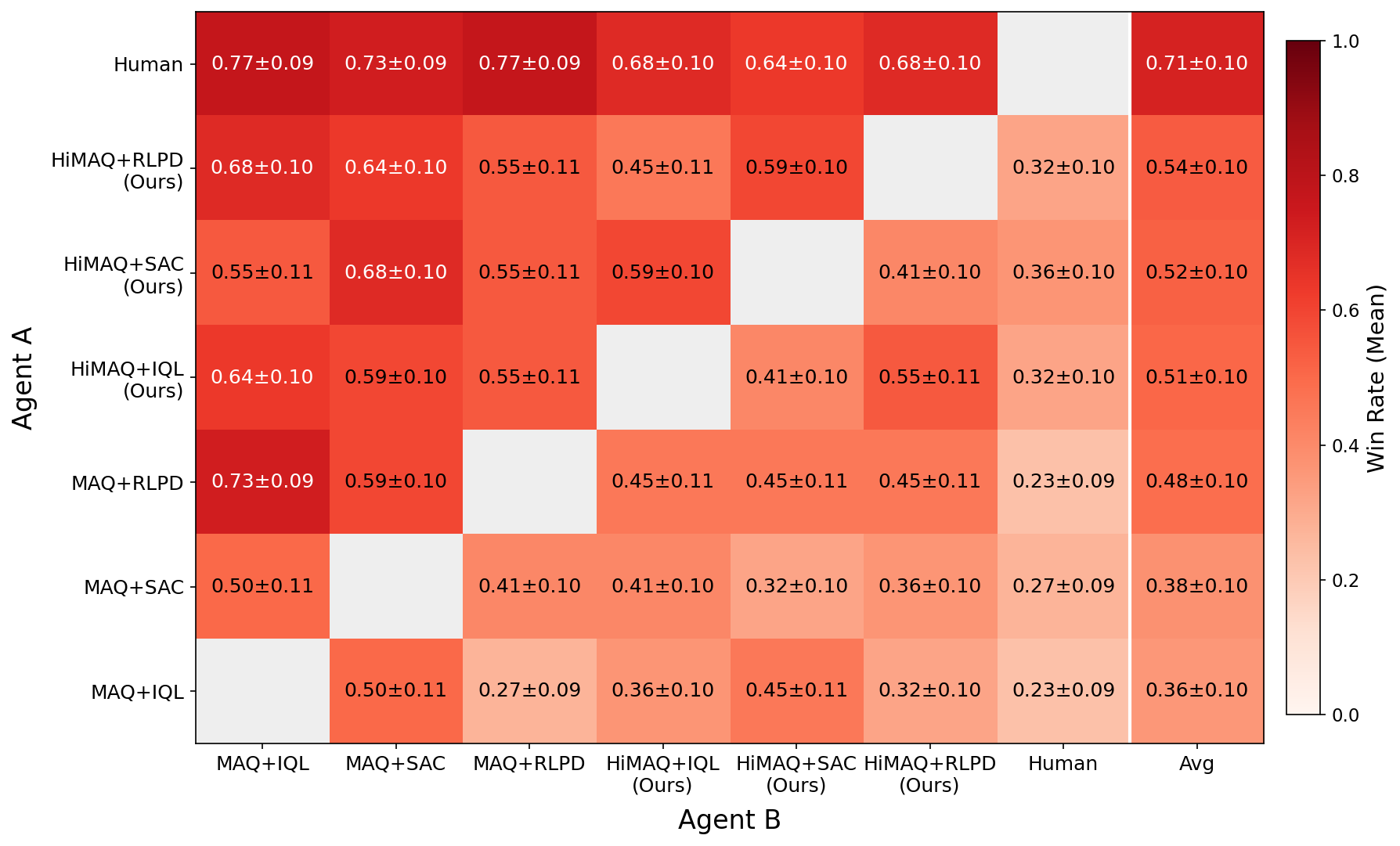}
    \caption{Human-likeness ranking test pairwise win rates.}
    \label{fig:human_total_survey_heatmap}
\end{figure}

While the Turing test compares agents against humans, the ranking test directly compares agents against each other.
Each evaluator is shown 21 pairs of videos, both from trained agents, and asked to choose which appears more human-like.
This produces pairwise win rates that reveal the relative human-likeness ordering among the human, MAQ, and HiMAQ agents in a single unified ranking. Fig.~\ref{fig:human_total_survey_heatmap} shows the pairwise win rates: each cell gives the fraction of trials in which Agent A (row) was judged more human-like than Agent B (column), with the \textit{Avg} column summarizing each agent's overall score.
The ranking by average win rate is:
Human (71\%) $>$ HiMAQ+RLPD (54\%) $>$ HiMAQ+SAC (52\%) $>$ HiMAQ+IQL (51\%) $>$ MAQ+RLPD (48\%) $>$ MAQ+SAC (38\%) $>$ MAQ+IQL (36\%).
All three HiMAQ variants consistently rank above their respective MAQ~\cite{guo2026learning} counterparts, and HiMAQ+RLPD is the closest to the human baseline, achieving a win rate of 54\% compared to the human's 71\%.

\subsubsection{Qualitative Feedback}
Evaluators' free-text comments help explain the rankings above.
RLPD-based agents were consistently preferred over their SAC- and IQL-based counterparts: SAC variants often failed to complete the task, immediately revealing them as non-human, while IQL variants contacted objects with a precision and force that felt mechanical; in contrast, RLPD's gradual, sweeping approach toward the goal was described as more natural.
Task difficulty also varied across environments, with \textit{Pen} the hardest to judge: for the in-hand reorientation it requires, evaluators often found no reliable distinguishing cue, and the strongest agents approach a near-chance Turing test win rate on this task, indicating their motion is nearly indistinguishable from human demonstration.
Interestingly, even a genuine human demonstration was sometimes judged non-human, e.g., one evaluator flagged a teleoperated \textit{Relocate} trajectory that overshot and corrected near the goal twice before settling, a teleoperation artifact rather than natural manipulation.

\section{Conclusion}
\label{sec:conclusion}

We present a human-like RL framework that produces action trajectories consistent with human behaviors while optimizing rewards. Our HiMAQ method hierarchically discretizes action sequences using a two-stage vector quantization hierarchy: the first stage discovers low-level subactions, while the second stage composes them into higher-level actions. Experiment results on D4RL demonstrate clear improvements over the non-hierarchical MAQ baseline, yielding more human-like behavior and better task performance. These benefits remain consistent across different RL algorithms such as IQL, SAC, and RLPD.

\noindent \textbf{Limitations.}
First, our evaluation is conducted entirely in simulation on four dexterous manipulation tasks from the D4RL Adroit benchmark~\cite{fu_d4rl_2021}, all sharing the same 24-DOF hand morphology. Without physical hardware, it remains unclear how well the learned human manifold transfers across real-world sim-to-real gaps or to structurally different domains such as locomotion.
Second, we assess human-likeness using trajectory-similarity proxies and a perceptual study with 22 evaluators. While these proxies do not fully capture human-like behavior, a larger and more diverse evaluator pool with formal significance testing would strengthen the findings.
Third, HiMAQ inherits limitations from its human demonstrations: the Adroit data are collected via teleoperation~\cite{rajeswaran2017learning}, introducing some jerkiness, and each task has only 25 trajectories, limiting motion diversity. Future work could incorporate smoothness regularization such as Lipschitz penalties~\cite{vuong2025action}.




\bibliographystyle{IEEEtran}
\bibliography{references}

\end{document}